# Dam Burst: A region-merging-based image segmentation method

Rui Tang[1], Wenlong Song[2], Xiaoping Guan[3], Huibin Ge[4] and Deke Kong[5]

[1]北京天地智绘科技有限公司
[234]China Institute of Water Resources and Hydropower Research (IWHR)
[5]北京治元景行科技有限公司

tangrui@tgis.top, songwl@iwhr.com, guanxiaoping@tgis.top, gehuibin@tgis.top, kongdeke@tgis.top

## Abstract

*Until now, all single level segmentation algorithms except CNN-based ones lead to over segmentation. And CNN-based segmentation algorithms have their own problems. To avoid over segmentation, multiple thresholds of criteria are adopted in region merging process to produce hierarchical segmentation results. However, there still has extreme over segmentation in the low level of the hierarchy, and outstanding tiny objects are merged to their large adjacencies in the high level of the hierarchy.*

*This paper proposes a region-merging-based image segmentation method that we call it Dam Burst. As a single level segmentation algorithm, this method avoids over segmentation and retains details by the same time. It is named because of that it simulates a flooding from underground destroys dams between water-pools. We treat edge detection results as strengthening structure of a dam if it is on the dam. To simulate a flooding from underground, regions are merged by ascending order of the average gradient inside the region.*

*The assessment program will be available at https://github.com/tgis-top/tGis/releases.*

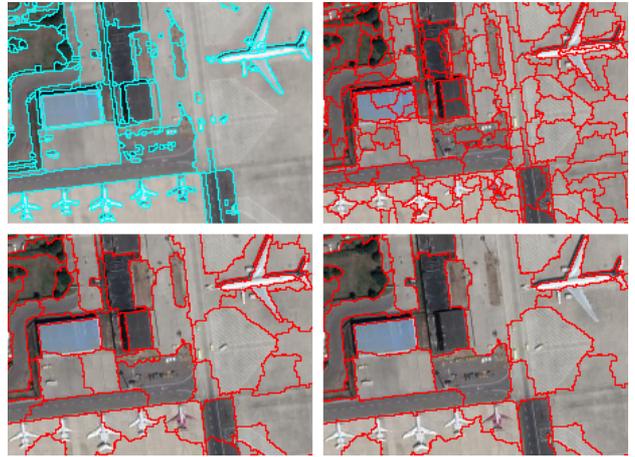

Figure 1: The proposed segmentation method (left top) avoids over segmentation and retains details by the same time. For hierarchical segmentation algorithms (Segmentation results courtesy of Zhongwen Hu et al.), over segmentation occurs in the low level of the hierarchy (right top), and outstanding tiny objects (plane et al.) are merged to their large adjacencies in the high level of the hierarchy (bottom).

## 1 Introduction

Image segmentation of high quality has significant improvement to object categorization, revealed by Andrew Rabinovich *et al.*. However, take image segmentation as pre-processing for object recognition is not wildly adopted. "The factor impeding the utility of segmentation for recognition is the unsatisfactory quality of image segmentation algorithms." Therefore, it is of great significance to improve the image segmentation algorithm.

Actually, in order to improve the effect of target recognition, we had turned to CNN for help, which hardly uses the results of image segmentation. Moreover, the segmentation itself has dominated by CNN. Nevertheless, "recent studies show that they are vulnerable to adversarial attacks in the form of subtle perturbations to inputs that lead a model to predict incorrect outputs". In addition, they lack explicability.

The key point of Graph-Based image segmentation methods are the determination of edge's weights. When dive into the deep origin we can conclude that the determination of edge's weights is the determination of difference between regions. Before shape works it is texture has the dominant effort. Texture is cyclically changes of intensity. Frequency and its location simultaneously form the pattern of texture. One cannot break the *Principle of Uncertainty* to determine frequency and its location at the same time.

Region growing, splitting and merging is according to how well each region fits some uniformity criterion. The uniformity criterion is also the determination of difference between regions, so it has the same defects as Graph-Based methods.

Active Contour Model, Level Set and Graph Cut are useful to extract single object, but not suitable for whole image segmentation. Watershed is too sensitive to local differences and noises, which lead to over segmentation.

Until now, all single level segmentation algorithms except CNN-based ones lead to over segmentation. And CNN-based segmentation algorithms have their own problems. To avoid over segmentation, multiple thresholds of criteria are adopted in region merging process to produce hierarchical segmentation results. The thresholds are related to specific criteria. When using different criteria, the thresholds are different.

Previously, Zhongwen Hu et al. had developed a general framework for analyzing and estimating the optimal thresholds for region merging. Earlier, they proposed a spatially-constrained color–texture model for hierarchical segmentation of very high resolution images. However, there still has extreme over segmentation in the low level of the hierarchy, and outstanding tiny objects are merged to their large adjacencies in the high level of the hierarchy.

Recently, we have come up with an idea of using edge detection results to constrain region merging. In fact, many other researchers also agree that edge detection results are the perfect cue for segmentation. And we designed a very novel mechanism to use edge detection results, which makes our method avoids over segmentation and retains details by the same time. When this method used as initial segmentation to produce hierarchy, it will make the hierarchy greatly simplified and make hierarchy simplification methods gain better results.

We also found an approximation way to calculate gradient by using haar-like box filter. This approximation can highly reduce gradient value inside the rough texture area that it avoids false edges to be extracted. Therefore, where will be less over segmentation.

## 2 Why over and under segmentation?

In the study of why need hierarchical segmentation results and why can't avoid over segmentation and retains detail in a single level result, we found that it is because of that one threshold for the whole image is never enough to distinguish all regions and inter-region heterogeneity less than intra-region heterogeneity is very common. To illustrate, we present two images here.

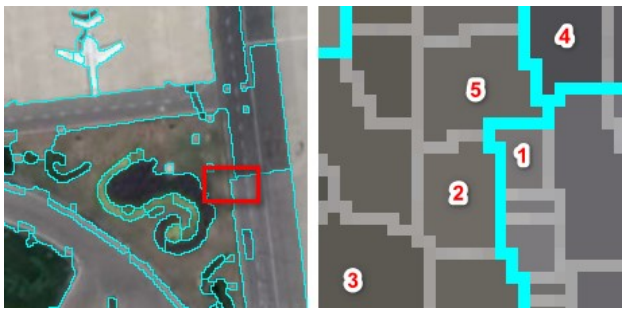

Figure 2: The Euclidean distance between block-2 and block-3 is greater than the distance between block-1 and block-2, and the distance between block-4 and block-5 is greater than the distance between block-1 and block-2.

The right image of figure 2 is an enlarged part form the red box in the left image. The cyan lines are final segmentation boundaries, and the gray lines are initial segmentation boundaries. Initial segmentation blocks are rendered by the mean intensity value inside the blocks. We choose 5 blocks which are marked with numbers to explain the two reasons.

The intensity values of the 5 blocks are *(120,117,116)*, *(112,108,101)*, *(91,88, 83)*, *(79,78,82)*, *(106, 102, 97)*.

### 2.1 Abnormal heterogeneity

It's easy to calculate that the Euclidean distance of intensity between block-1 and block-2 is *19.24* and the Euclidean distance between block-2 and block-3 is *34.13*. From a human point of seeing, block-2 and block-3 should be merged, and block-1 and block-2 should not. However the numerical difference between block-2 and block-3 is greater. A threshold greater than *19.24* will let block-1 and block-2 merged, but keeps block-2 and block-3 separate, which are under-segmented for the driveway in right bottom of figure 2. A threshold less than *19.24* will lead to over segmentation of the planted area in left bottom.

Usually, we think inter-region heterogeneity is greater than intra-region heterogeneity and intra-region homogeneity is greater than inter-region homogeneity. When this is not the case, we call it abnormal heterogeneity. However, as you can see from the calculations described above, inter-region heterogeneity less than intra-region heterogeneity is very common, which will cause many algorithms to fail.

### 2.2 One threshold is not enough

We can also calculate the Euclidean distance between block-4 and block-5, which turn out to be *39.12*. And we already know the Euclidean distance of intensity between block-1 and block-2 is *19.24*. This indicates that a threshold slight less than 39.12 might be ok to separate the driveway in right up and the planted area in left bottom, but this threshold will definitely make the planted area in left bottom merged with the driveway in right bottom.

The above analysis shows that to make good segmentation different thresholds should be used for different regions.

## 3 Our solutions

The direct idea to solve the problem is to apply different thresholds to different regions. But it is very difficult to determine the appropriate thresholds. We found two ways to solve them indirectly.

### 3.1 Make use of hysteresis of Canny Edge Detector

First, we know that gradient tell the difference between two sides of a pixel, to some extent gradient is an inter-region heterogeneity measurement, and edge detection results are the best divides of regions. Dual threshold and connectivity constraint of Canny Edge Detector make edge detection results contain a large range of gradient value, which can be considered as different thresholds are used for different regions.

Edge detection results are usually not closed. So we try to connect the breaks of edges to make good segmentation. When edge detection results and fine-grained segmentation results are displayed together in layers, we found that boundaries of fine-grained segmentation will connect edges properly. Especially, Canny Edge Detector and Watershed segmentation results are of high coincidence. Therefore, we adopt Watershed as initial segmentation method. Moreover, they all need gradients.

Apparently, after removing boundaries with no edge detection results (edge pixels) on them, the boundaries left behind will make a very good segmentation. But this sample rule is not always correct; when not, wrong segmentation occurs. We found an exhilarating method to remove useless boundaries. The method has some similarity to Watershed. They all proceed by ascending order of elevation (gradient). Watershed builds "Dams", our method destroys "Dams", and so we give it a metaphorical name: Dam Burst.

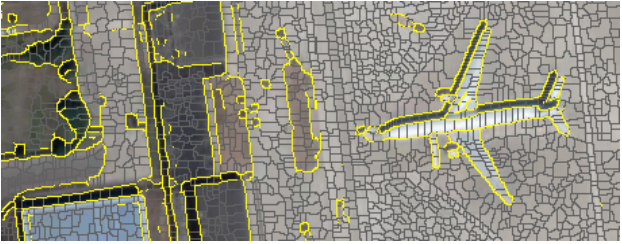

Figure 3: Canny edge detection results and Watershed segmentation results are displayed together in layers.

### 3.2 Suppress gradient inside the rough area

Since texture is cyclically changes of intensity, average of intensity in a texture cycle is always same. Mostly, averages in different texture cycles are different. When the value of the cyclically change is large, the area is rough, that is, the region has a high degree of intra-region heterogeneity.

When calculating the gradient of a pixel with the mean intensity of a certain area on both sides, the gradient in the rough area is suppressed. Therefore, fewer false edges will be detected. In turn, over segmentation in rough areas can be reduced.

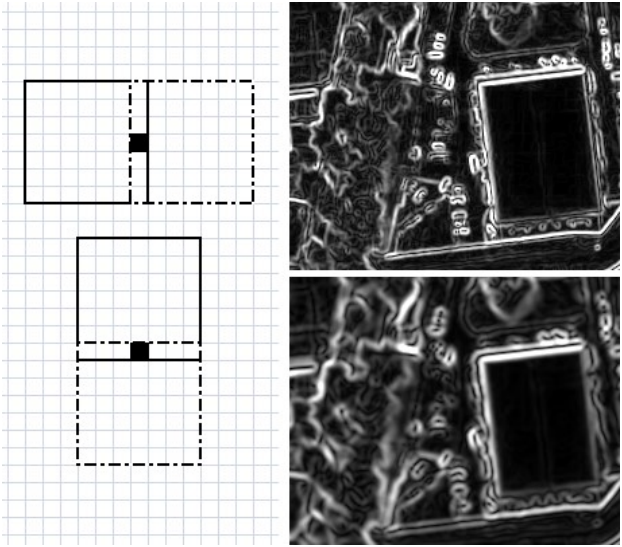

Figure 4: Gradient at the tiny dark cell (pixel) is computed by the difference of averages from solid border box area and dashed border box area. Compared to the Sobel gradient (right top), the Haar-like gradient (right bottom) in the rough area is suppressed and its local maximum value is somewhat misplaced.

In order to speed up gradient calculations with an integral image, we use the two placements of boxes like figure 4 to calculate horizontal and vertical gradient separately. Tiny square cell represents pixel, and the gradient at solid black ones are calculated by the difference of average intensity from solid border box area and dashed border box area. The width of the boxes should be equal to the maximum of min cycle of all textures in image. The gradient error caused by imprecise equality of given cycle and real cycle is suppressed by averaging. Thereby, do not bothering with the value of the box width.

After experiments, we found that taking value of 5, 7, 9, 11, 13 and 15 for the box width can satisfy most of the cases. Smaller values can cause insufficient suppression, and larger values can cause local maximum to be misplaced severely.

## 4 Details of Dam Burst

As mentioned before, boundaries of initial segmentation with no edge pixels on them are not always suitable for removing. The key point of Dam Burst is to make the boundaries (dams) with no edge pixels on them become always suitable for removing. We use a Watershed-like approach to ensure this. Besides, the defects of edge detector leave breaks in edges. Those breaks lead to wrong combination of water-pools. We designed a coarse method to locate dam at the breaks of edges directly, then apply an inter-region heterogeneity measurement for further confirmation.

### 4.1 Main order to remove dam

Denoting the gradient image as $I_g$, the *merging order* of region (water-pool) $R_i$ is given by

$$MO(R_i) = \text{avg}\left(I_g(p|p \text{ in } I_g \cap R_i)\right). \quad (1)$$

Where $I_g(p)$ is a function that fetches the gradient value from $I_g$ at position p. $MO(R_i)$ means the average gradient value inside water-pool $R_i$.

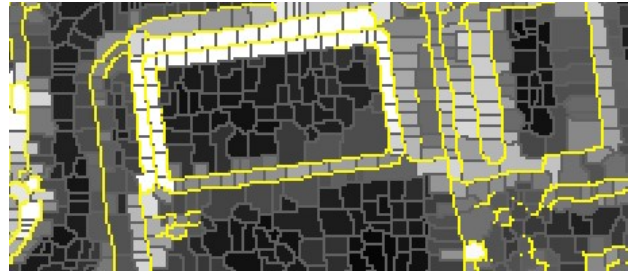

Figure 5: Water-pools filled with gray value that is in proportion to $MO(R)$. Yellow lines are edge detections results.

From figure 5, we can see that the darkest pools are at the center of the desired segmentation, as it gets to the border the pools are get brighter. Intuitively, we might think that let the pool that has the smallest $MO(R)$ of all its neighbors be the seed and then make them growing. Yes, this works. Because, if two regions are not supposed to merge, when they grow to meet, it is of high possibility that there are edge pixels on their adjacent boundary. Therefore, it makes the boundary not suitable for removing, that is, the dam is not suitable to destroy.

Like Watershed, our method simulates a flooding from underground to let region grow. So before removing dams, sort all water-pools by ascending order of MO(R). For a water-pool $R_i$, the dam between it and its adjacent neighbor $R_j$ can be broken only if $R_j$ has a smaller $MO(R_j)$. Surely there are more conditions need to meet which will be explained later.

At the upside of figure 5, we can see some pools that have biggest MO(R) are also at the center of the expected segmentation. Fortunately, there do will be some edge pixels on at least one boundary of those pools. To let edge detection results help, we introduced the concept of *strength of dam*.

We treat the edge detection result as strengthening structure of a dam if it is on the dam. Let the edge detection result be $I_e$, and denote the dam between $R_i$ and $R_j$ as $E_{ij}$. Then the strength of the dam $C_{ij}$ is defined as follow,

$$C_{ij} = \frac{\text{Len}(E_{ij} \cap I_e)}{\text{Len}(E_{ij})}. \quad (2)$$

Len(∗) represents a function that measures the length. Then, $C_{ij}$ means the ratio of the length of strengthening structure to the length of the dam.

We use a threshold value $T_c$ to determine whether a dam is strong enough to be not removed. The recommended range of $T_c$ is in (0,0.5). The dam $E_{ij}$ cannot be removed only if $C_{ij}$ greater than $T_c$. We call a dam *weak dam* when strength of it is smaller than $T_c$.

### 4.2 Locate dam at the breaks of edges

As can be seen from figure 5, if the water-pool is in the center of the region, there is no edge pixels on any of its dams. We already know that regions are merged in ascending order of the average gradient inside regions. When the water-pool grows to reach the border of the region, it will meet the breaks of edges. Therefore, when a dam is at the breaks of edges, there must be lots of edge pixels on at least one other dam that connected with the dam, that is, there is a water-pool that contains the dam has some strong dams.

So we introduced *strength index* of the entire water-pool and used a threshold value $T_{rsi}$ to determine whether a water-pool has dams at the breaks of edges. *Strength index* of $R_j$ is give by

$$RSI(R_j) = \frac{\Sigma_{C_{ij} > T_c} C_{ij}}{\Sigma_{C_{ij} > T_c} 1}. \quad (3)$$

Where *i* indicates all the adjacent neighbor of $R_j$. So, *strength index* is the average strength of a pool's dams with the strength greater than $T_c$.

When $RSI(R_j) > T_{rsi}$ means $R_j$ is strong water-pool and the weak dams of $R_j$ may be at the breaks of edges.

The value of $T_{rsi}$ can be a little bit bigger than $T_c$, equal to $T_c$, or smaller than $T_c$. When $T_{rsi}$ is smaller than $T_c$, it means R is strong water-pool if RSI(R) is greater than zero.

### 4.3 Apply inter-region heterogeneity measurement

If the two water-pools are of high inter-region heterogeneity, the possibility of the dam between them being at the breaks of edges is increased. So we add an inter-region heterogeneity measurement to locate dams at the breaks of edges.

For a water-pool $R_i$ and its adjacent neighbor $R_j$, let their inter-region heterogeneity be $InD_{ij}$. Give a threshold $T_{ind}$, in the process of Dam Burst, if $R_i$ is strong (that is to say $RSI(R_i) > T_{rsi}$), the dam between $R_i$ and $R_j$ can be broken only if the additional condition that $InD_{ij} \leq T_{ind}$ is met; if $R_i$ is weak, the dam removing conditions stay unchanged.

Unlike $T_c$ and $T_{rsi}$, $T_{ind}$ is auto estimated in the process of Dam Burst. Init $T_{ind}$ with zero, and in the process of Dam Burst, update $T_{ind}$ by

$$T_{ind} = \max\left(T_{ind}, \frac{\Sigma_r InD_{ij}}{\Sigma_r 1}\right). \quad (4)$$

That is, update $T_{ind}$ to the greater one of itself and the average inter-region heterogeneity of removed water-pools. Why update $T_{ind}$ to the average? Because when no edge pixels on a region's boundary it does not mean there really no, defects of edge detection will contribute. Therefore, an improvement of this method could lay on edge detection.

---

**Algorithm 1** Dam Burst
---
**Input**: RAG of initial segmentation
**Parameter**: $T_c$, $T_{rsi}$
**Output**: Merged RAG
1: Let $T_{ind} = 0$ and Merged = true.
2: **while** Merged **do**
3:     Sort all water-pools by ascending order of MO(R)
4:     Let $C_{ind} = 0$, $\Sigma_{ind} = 0$ and Merged = false
5:     **for each** water-pool ($R_a$) in sorted water-pools
6:         Let $InD_{min}$ = DBL_MAX and $R_m$ = NULL
7:         **for each** adjacent water-pool ($R_b$) **of** $R_a$
8:             **if** $MO(R_b) > MO(R_a)$ or $C_{ab} > T_c$
9:                 continue
10:            end if
11:            **if** $RSI(R_a) \leq T_{rsi}$
12:                 **if** $InD_{ab} < InD_{min}$
13:                     Let $InD_{min} = InD_{ab}$ and $R_m = R_b$
14:                 end if
15:            **else if** $InD_{ab} \leq T_{ind}$
16:                 **if** $InD_{ab} < InD_{min}$
17:                     Let $InD_{min} = InD_{ab}$ and $R_m = R_b$
18:                 end if
19:            end if
20:         end for each
21:         **if** $R_m$ != NULL
22:            merge $R_a$ and $R_m$
23:            Let Merged = true
24:            Let $C_{ind} += 1$, $\Sigma_{ind} += InD_{min}$
25:            Let $T_{ind} = \max(T_{ind}, \Sigma_{ind}/C_{ind})$
26:         end if
27:     end for each
28: **end while**
29: **return** water-pools (Merged RAG)

---

Specific inter-region heterogeneity measurement can be determined according to the characteristics of the image. For the sake of convenience, we adopted Euclidean distance of mean intensity inside the water-pool.

### 4.4 Order to remove a pool's dams

There may be more than one water-pool around $R_i$ meet the conditions that has a smaller MO(R) and $C_{ij} \leq T_c$. Which is the exact one that the dam between it and $R_i$ can be removed? Or should all of them merged together? There are no differences when apply to orthophoto and photo with short depth of field. Otherwise, only removing the most proper dam will gain better results.

Compared with gradient, $InD_{ij}$ tell the difference between regions more directly and precisely, thus we chose the dam between $R_i$ and $R_j$ with smallest $InD_{ij}$ to remove.

### 4.5 Procedures of Dam Burst

After two water-pools combined, some dams will get longer, but the strengthening structures are still the same. So, after an iteration of Dam Burst, there will be some new weak dams appear. Therefore, we should run Dam Burst several times until there are no new weak dams emerge. The procedures of the proposed method are given in Algorithm 1.

## 5 Experiments

Experiments and comparisons were carried out to validate the effectiveness of the proposed method in reducing over segmentation. Berkeley Segmentation Data Set 500 (BSDS500), some aerial images and some satellite images are used. Firstly, we explored the effect of parameters on the proposed method, and then compared it with a state-of-art hierarchical segmentation algorithm.

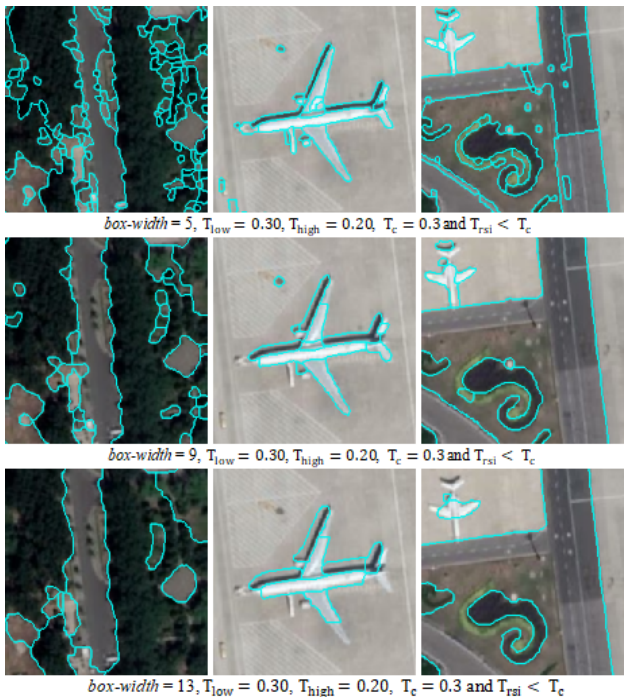

Figure 6: Analysis of *box-width*. From top to bottom, *box-width* is 5, 9 and 13. All other parameters are fixed, where $T_{low} = 0.30$, $T_{high} = 0.20$, $T_c = 0.3$ and $T_{rsi} < T_c$.

### 5.1 Analysis of Different Parameters

Our algorithm introduces only three new parameters, *box-width*, $T_c$ and $T_{rsi}$. Their values are very intuitive. The high threshold (denoted as $T_{high}$) and low threshold (denoted as $T_{low}$) of Canny Edge Detector are determined by the percentage of pixels after NMS that need to be retained.

In order to analyze the effects of the *box-width*, we fixed other parameters and did three sets of experiments. As show in figure 6, *box-width* is 5, 9 and 13 from top to bottom. Obviously, the first row of images is a little over-segmented, and the next two rows of image show a gradually serious under segmentation. Because when *box-width* gets larger, the pixels after NMS get fewer. The same $T_{high}$ and $T_{low}$ results in a significant reduction in edge pixels.

In addition, the boundaries of the small targets in the bottom two rows of image gradually deviate from the correct position. So it doesn't make much sense to give too larger value for *box-width*. The recommended values of $T_{high}$ and $T_{low}$ corresponding to *box-width* are shown in Table 1.

| Box-width | $T_{high}$ | $T_{low}$ |
|---|---|---|
| 5 | 0.210 | 0.300 |
| 7 | 0.250 | 0.370 |
| 9 | 0.290 | 0.450 |
| 11 | 0.300 | 0.470 |
| 13 | 0.307 | 0.490 |
| 15 | 0.315 | 0.500 |

Table 1: recommended values of parameters

From the top row of figure 7, we can see that using larger $T_{high}$ and $T_{low}$ do reduce under segmentation. And as $T_c$ is increased, the under segmentation occurs again, which can be seen from bottom row of Figure 7.

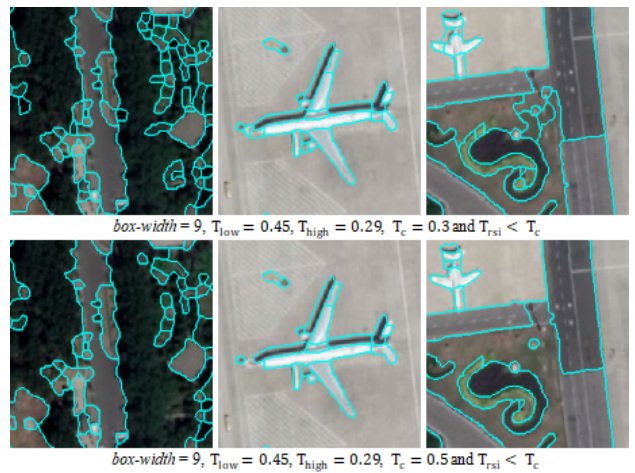

Figure 7: Analysis of $T_{low}$, $T_{high}$ and $T_c$. When a larger *box-width* is provided, larger $T_{low}$ and $T_{high}$ are required to reduce under segmentation. Increasing $T_c$ also results in under segmentation.

The left column of figure 6 and figure 7 is a small wood, and its texture is very rough. The area not covered by leaves is earthy. Some of the leaves are light, some are dark, and the shadow is black. The whole area is finely filled with these colors. So this image is prone to over segmentation. We can get a good segmentation by giving a slight larger *box-width*. The texture of middle image is relatively simple, and it is easy to find parameters that produce good segmentation.

The texture of the planted area in right image is very rough too. It is more difficult to make a good segmentation, and many parameters cause them to merge with the adjacent driveway. The reason is that its color block granularity is too large. In this case, we can't just increase *box-width* to avoid over segmentation. Besides, too large B will lead to the deviation of the segmentation boundary. It seems that it is better to process rough and flat areas separately, which can be another improvement idea of this algorithm.

### 5.2 Comparisons with a state-of-art method

Although this is single-level image segmentation algorithm, we still compare it to a state-of-art hierarchical algorithm. Because this algorithm outperforms many other single-level image segmentation algorithms too much, Watershed, N-Cut, Region Growing and others, there is no need to put them together for comparison. Furthermore, these single-level image segmentation methods only serve as initial segmentation step for this method and other hierarchical algorithm.

As far as my assessment is concerned, the hierarchical image segmentation method from Zhongwen Hu et al. is one of the best, and they provided executable program.

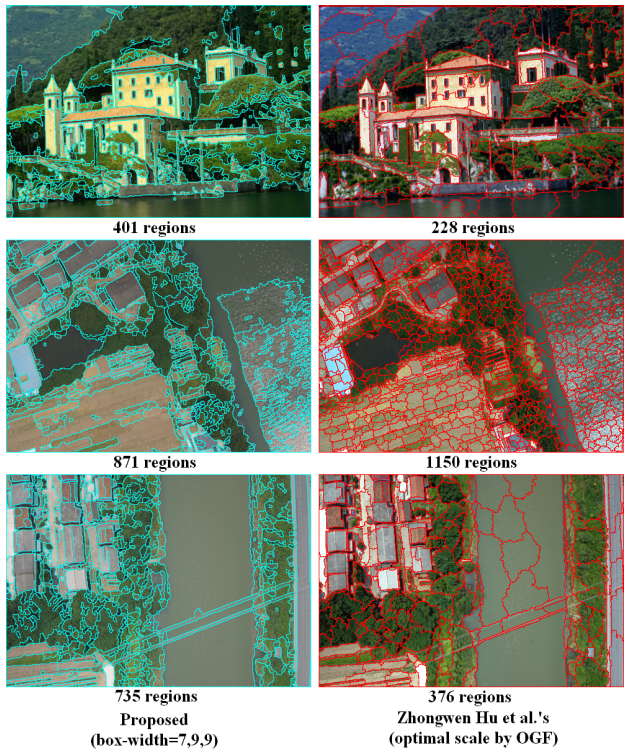

Figure 8: Comparison with state-of-the-art simplification methods. The proposed method reduces over segmentation greatly.

The parameters in Table 1 were used in the comparison. It can be seen from the comparison that this algorithm has a great improvement in reducing over segmentation. If this algorithm is used as the initial segmentation method of hierarchical segmentation algorithm, the hierarchies will be greatly simplified.

Nevertheless, there is still over segmentation. Because of under segmentation will result in more serious negative influences than over segmentation for later image interpretation, parameters are chosen to completely avoid under segmentation. Actually, most of the tiny regions do have significant difference with their surroundings. When these small regions are merged into their large neighbors, it will lead to unreasonable changes in the inter-region heterogeneity.

## 6 Conclusion

In this paper, we propose a new region-merging-based image segmentation method that we call it Dam Burst, and present a new way to compute gradient that can suppress gradient inside the rough area. The segmentation method proposed here outperforms many previously proposed non-CNN-based single level method. When compared with CNN-based methods, this method is more explainable, more generalizable and simpler.

Using one uniformity criterion to segment every corner of an image is apparently not good enough. That is why lots of segmentation methods produce hierarchical results, and different categories get the best results at different levels. Thanks to the hysteresis of Canny Edge Detector and the merging order of Dam Burst, all categories are well segmented in one single-level result by our method; thereby our method keeps details and avoids over segmentation by the same time.

The hysteresis of Dam Burst also appears in the evaluation of threshold of inter-region heterogeneity. The threshold increases as the procedures going on. That leads different threshold for different regions of image.

The most important part of this work is that it's also a framework for segmentation. The method to detect edges and the measurement of inter-region heterogeneity can be replaced with more powerful ones to get better result. In the future, we will further study these aspects, and make meaningful improvement to this method.

### Acknowledgments

The authors would like to thank *Beijing ReAvenue Technology Co., Ltd* for their continuous trust and support for non-deep-learning computer vision algorithms, thank *Prof. Zhongwen Hu* for providing evaluating software, and thank the anonymous reviewers for their highly thought-provoking remarks.